The paper was accepted in IEEE Journal of Biomedical and Health Informatics 1

# Click, Predict, Trust: Clinician-Guided and Clinician-Evaluated AI Segmentation for Lung Cancer CT-Based Prognosis within the Knowledge-to-Action Framework

Mohammad R. Salmanpour, Sonya Falahati, Amir Hossein Pouria, Amin Mousavi, Somayeh Sadat Mehrnia, Morteza Alizadeh, Arman Gorji, Zeinab Farsangi, Alireza Safarian, Mehdi Maghsudi, Carlos Uribe, Arman Rahmim, Ren Yuan

***Abstract*—Lung cancer is the leading cause of cancer death, with CT central to screening, prognosis, and treatment. Manual segmentation is variable and time-consuming, while deep learning (DL) offers automation but faces adoption barriers. Guided by the Knowledge-to-Action-framework, this study develops a clinician-guided and clinician-evaluated pipeline to improve reproducibility, prediction, and clinical trust. Multi-center lung-cancer CT from 999 patients across 12 public datasets were analyzed. Five DL models (3D Attention U-Net, ResUNet, VNet, ReconNet, SAM-Med3D) were benchmarked against expert contours on whole and cropped images. Segmentation reproducibility was tested with 497 PySERA-extracted radiomics features using Spearman correlation, ICC, Wilcoxon tests, and MANOVA. Prognostic modeling compared supervised (SL) and semi-supervised learning (SSL) with 38 dimensionality reduction strategies and 24 classifiers. Six physicians qualitatively evaluated masks across 21 questions spanning seven domains, including clinical meaningfulness, boundary quality, survival prediction, trust, and workflow integration. VNet-based segmentations achieved the highest performance (Dice 0.83±0.07, IoU 0.71±0.09), strong radiomics reproducibility (average Spearman correlation coefficient 0.76, ICC 0.65), and strong predictive SSL performance (accuracy 0.88±0.003, F1=0.83±0.004). SSL generally outperformed SL across models. Descriptive physician ratings indicated favorable boundary quality and peritumoral representation for VNet, although Friedman tests showed no statistically significant differences in selected physician-rated domains. Importantly, experts preferred AI to provide initial masks for refinement rather than replace physicians. This study shows that VNet, integrated with SSL, can deliver accurate, reproducible, and potentially clinically useful tools for CT-based lung cancer prognosis. Embedding AI as a supportive assistant within physician-centered workflows offers a feasible path to clinical translation.**

## I. INTRODUCTION

For artificial intelligence (AI) in medical imaging to achieve clinical adoption, technical advances must be embedded within frameworks that address barriers to implementation [1]. The Knowledge-to-Action (KTA) framework within implementation science [2] provides such a roadmap, emphasizing that innovations must extend beyond methodological performance to overcome systemic, clinical, and workflow challenges. Previous studies [3], [4], [5], [6], [7], [8] have highlighted the significant barriers to AI adoption in healthcare, including infrastructure issues, clinical skepticism, and the need for human-centered decision-making in AI applications. These findings align with the KTA framework's assertion that AI innovations must address both technical and

This manuscript was submitted for review on January 13, 2026, by Mohammad R. Salmanpour. This work was supported by the Natural Sciences and Engineering Research Council of Canada (NSERC) Discovery Horizons Grant DH-2025-00119, as well as the UBC Department of Radiology 2023 AI Fund. (Corresponding author: Mohammad R. Salmanpour). Arman Rahmim and Ren Yuan are credited as co-last authors for their equal contributions to this study.

Mohammad R. Salmanpour, Arman Rahmim, and Carlos Uribe are with the Department of Basic and Translational Research, BC Cancer Research Institute, Vancouver, BC, Canada (e-mail: msalman@bccrc.ca, arman.rahmim@ubc.ca, and curibe@bccrc.ca).

Mohammad R. Salmanpour, Arman Rahmim, and Carlos Uribe are with the Department of Radiology, University of British Columbia, Vancouver, BC, Canada (e-mail: msalman@bccrc.ca, arman.rahmim@ubc.ca, and curibe@bccrc.ca).

Ren Yuan is with BC Cancer, Vancouver Center, Vancouver, BC, Canada (email: ren.yuan@bccancer.bc.ca).

Mohammad R. Salmanpour, Sonya Falahati, Arman Gorji, and Mehdi Maghsudi are with the Technological Virtual Collaboration (TECVICO Corp.), Vancouver, BC, Canada (e-mail: msalman@bccrc.ca, falahati.sonya@gmail.com, gorji.arman@edu.umsha.ac.ir, and dr.mmaghsudi@gmail.com).

Amir Hossein Pouria is with the Department of Computer Engineering, Amirkabir University of Technology, Tehran, Iran (e-mail: amirhossein.pourya@aut.ac.ir).

Amin Mousavi is with the Department of Computer, Abhar Branch, Islamic Azad University, Abhar, Iran (e-mail: amin.mousavi.academic@gmail.com).

Somayeh Sadat Mehrnia is with the Department of Radiology Technology, Faculty of Allied Medical Sciences, Shahid Beheshti University of Medical Sciences, Tehran, Iran (e-mail: s.mehrnia95@gmail.com).

Morteza Alizadeh is with the Department of Mathematics, University of Isfahan, Isfahan, Iran (e-mail: alizadehmorteza2020@gmail.com).

Zeinab Farsangi and Alireza Safarian are with the Rajaie Cardiovascular Medical and Research Center, Iran University of Medical Science, Tehran, Iran (e-mail: zeinabfarsangi@yahoo.com and a.safarian@salk.at).

The complete workflows—including the full codebase are available on GitHub via https://github.com/MohammadRSalmanpour/Clinician-in-the-Loop-AI-Segmentation-for-CT-Based-Prognosis.

Informed consent was obtained from all patients in the publicly available datasets, all of which are appropriately cited in the manuscript.

systemic challenges. Our study further supports these conclusions by demonstrating how the KTA framework can guide AI integration, emphasizing the importance of addressing clinical workflows, reproducibility, and prediction performance in real-world settings. Our own prior work [2], specifically identified 11 barriers to AI diagnostic decision support tools adoption in diagnostic medicine: (i) infrastructure and integration, (ii) clinical skepticism and loss of human-centered decision-making, (iii) insufficient adaptability to clinical complexity, (iv) trust and accountability in AI-generated reporting, (v) ethical, legal, financial, and privacy concerns, (vi) lack of training and familiarity, (vii) workflow disruption and lack of standardization, (viii) workforce resistance due to job concerns, (ix) limitations in atypical or complex cases, (x) clinical distrust in AI decision-making, and (xi) bias, reliability, and quality challenges. Addressing these barriers is essential if AI is to move from promising prototypes to sustainable, evidence-based tools.

Lung cancer remains the leading cause of cancer-related mortality worldwide (>1.8 million deaths annually) [9], with ~124,000 deaths projected in the United States by 2025 [10]. Despite therapeutic advances, prognosis remains poor, underscoring the need for earlier detection, better prognostic assessment, and imaging-driven innovation—gaps aligned with KTA barriers of adaptability (iii) and workflow disruption (vii) [11], [12]. Large randomized trials—the U.S. National Lung Screening Trial (NLST) and the Dutch–Belgian NELSON trial—established low-dose CT as the only screening modality proven to reduce lung cancer mortality [13], [14]. Because CT is high-resolution, widely available, and relatively low cost, it is central to screening and follow-up, addressing infrastructure (i), workflow (vii), and clinical skepticism (ii) while mitigating financial concerns (v) [15]. Although PET/CT remains essential for staging [16], CT is the natural platform for embedding AI-enabled radiomics and prognostic pipelines [17], [18]. Ethical, legal, and privacy considerations (v) remain outside the scope of this work.

Segmentation remains a major bottleneck in quantitative imaging, as accurate delineation of tumor boundaries directly determines the reliability of radiomic features and downstream prognostic modeling [18], [19]. Manual segmentation is subjective, time-intensive, and variable, with inter-observer differences often exceeding the differences of biologically relevant changes [20]. Fully automated deep learning (DL) models promise efficiency but often fail to generalize across scanners, institutions, and tumor morphologies [21], [22]. Radiomic features become unstable, prognostic models irreproducible, and trust erodes [23]. Here, KTA barriers of adaptability (iii), trust (iv), clinical distrust (x), and reliability (xi) converge. Variability arises from observer differences, architectural diversity, irregular tumors, and multi-center heterogeneity [24], [25]. Left unresolved, these errors propagate into radiomics and outcome models [26]. Most prior studies evaluated segmentation performance solely through geometric metrics such as Dice or IoU, which quantify overlap but not whether contours preserved radiomics stability or prognostic value [27], and limiting translation [28].

One practical solution is tumor-centered cropping with a peritumoral margin [29], which not only standardizes inputs and reduces background noise, but also captures the radiologically meaningful peritumoral features that are known to have prognostic information. Technically, cropping lowers input size and memory demands, improving computational efficiency and lowering costs. Clinically, it enhances reproducibility across heterogeneous imaging. In KTA terms, cropping addresses infrastructure (i), adaptability (iii), workflow disruption (vii), and reliability (xi) [30]. Building on this, the study proposes a multi-pronged solution: (i) semi-automated segmentation with click-point or bounding-box guidance aligns physician oversight with algorithmic predictions, addressing workflow (vii) and skepticism(ii) [31], [32]; benchmarking five DL architectures across multi-center datasets evaluates reproducibility, addressing reliability (xi) and adaptability (iii) [30], [33]; radiomics stability analyses ensure features remain robust, targeting trust (iv) and clinical distrust (x) [24]; and (iv) expert qualitative review validates plausibility, countering skepticism (ii) and workforce resistance (viii) by showing AI supports rather than replaces clinicians [34].

Moving beyond Dice and IoU, which primarily assess the geometric fidelity of segmentation algorithms, radiomics analyses offer a radiologically meaningful evaluation [27]. They capture not only lesion size and shape but also intra-lesional textural heterogeneity, a feature highly sensitive to segmentation accuracy [35]. Outcome prediction further connects DL-based radiomics to patient survival, testing whether the generated contours encompass prognostically relevant regions [18], [36], [37]. By preserving radiomic features, such approaches retain subtle, prognostically informative signals; in atypical cases (ix), DL models can demonstrate that maintaining these features enhances outcome prediction, thereby strengthening clinical trust [38]. Still, prediction tasks face a scarcity of annotated outcome datasets [39]. Semi-supervised learning (SSL) addresses this by leveraging unlabeled data to improve generalization across multi-center cohorts [40], [41], lowering annotation burdens and addressing adaptability (iii) and lack of training/familiarity (vi) [42]. Previous studies [40], [43], [44] have shown that SSL models can be more generalizable than conventional supervised learning (SL) models when labeled data are limited, as they can leverage additional unlabeled data during training. In the present study, SSL was therefore included as a complementary downstream validation framework to investigate whether radiomics features derived from DL-generated segmentation masks remain useful for prognostic modeling under limited outcome-label availability and whether their predictive performance generalizes across heterogeneous datasets.

Quantitative radiomics metrics must be complemented by clinician-guided and clinician-evaluated assessment [42], [45] validating clinical plausibility and reinforcing interpretability [34]. Clinicians contributed throughout the study through lesion delineation, ground-truth verification, evaluation of AI-generated masks, and interpretation of their clinical and prognostic relevance. In addition, the 11 KTA barriers guiding this framework were identified in the previous study through input from clinical professionals [2]. A key principle is that AI should augment, not replace, clinical expertise [46]. DL can streamline segmentation, reduce variability, and enhance reproducibility, but physicians remain essential in oversight and decision-making [30]. Positioning AI as a clinical assistant addresses workforce resistance (viii) and strengthens trust (x) [32]. By showing how DL highlights subtle features while radiologists refine contours, this study reinforces the centrality of physician oversight [47]. Most prior work [23], [27], [48], [49] stopped at geometric evaluation, rarely linking segmentation variability to radiomics reproducibility and survival, or embedding physician oversight. Few [29], [50] tested cropping, outcome-guided validation, or physician-guided evaluation review. The absence of these steps explains why DL segmentation, despite strong technical performance, has struggled with adoption [51], [52]. In the current study, we aim to fill that gap by integrating cropping, benchmarking, radiomics stability, outcome prediction, and physician-guided evaluation review within a single KTA-aligned framework. In doing so, it could address nearly all barriers to adoption, providing a roadmap for trustworthy AI in lung cancer imaging.

In summary, the current study makes four contributions (outlined next) that are aligned with KTA, tackling 10 of 11 above-mentioned barriers to adoption of AI: First, benchmarking five DL architectures for lung lesion segmentation across multi-center CT datasets tackles reliability (xi), adaptability (iii), and infrastructure (i). Second, radiomics stability analyses move beyond geometric accuracy to confirm prognostic preservation, aiming to better address clinical distrust (x) and trust/accountability (iv). Third, a semi-automated, clinician-guided/evaluated SSL framework reduces workflow disruption (vii), mitigates skepticism (ii), and lowers annotation burdens tied to training and familiarity (vi). Fourth, comparing ground truth- and DL-based radiomics features in ML classification with physician review explains when DL performs better or worse, addressing variability (xi), atypical cases (ix), and workforce resistance (viii) by confirming AI's role as assistant, not replacement. While this study tackles nearly all KTA barriers, ethical, legal, financial, and privacy concerns (v) remain beyond the scope and require institutional and policy solutions. Collectively, these contributions demonstrate not only technical feasibility but also a barrier-focused pathway for embedding AI into clinical practice.

## II. METHOD AND MATERIAL

### *A. Patient/Imaging Data.*

Clinical data, CT scans, and manually delineated lesion masks were obtained from 999 patients across 12 TCIA datasets, as summarized in Table I. The included cohorts were selected to provide heterogeneous, multicenter data spanning diverse patient demographics, tumor characteristics, imaging parameters, and clinical outcomes. Available demographic and clinical characteristics varied across datasets and are reported in Table I. For prognostic analysis, patients with available overall survival (OS) data were categorized as Class 1 (survival >4 years; favorable prognosis) and Class 0 (survival <4 years; poor prognosis). By integrating 12 heterogeneous, multi-center datasets with diverse demographics, tumor subtypes, and outcome profiles, this study directly addresses KTA barriers of insufficient adaptability to clinical complexity (iii), bias and reliability challenges (xi), and infrastructure and integration (i) by ensuring that the developed framework is tested across varied real-world conditions rather than narrow, single-center cohorts.

### *B. Mask Delineation and Expert Verification.*

Chest CT scans were first assessed in the standard lung window with multiplanar reconstructions to identify nodules displaying features suspicious for malignancy, such as irregular morphology, spiculated or lobulated margins, abnormal attenuation, or rapid growth. Lesions deemed suspicious were manually segmented on each axial slice in 3D Slicer (v5.8) by two board-certified diagnostic physicians (thoracic radiology) working in consensus, adhering to a prespecified image-based lung-tumor contouring protocol; pathology/MTB, when available, was used only to confirm lesion identity and did not alter boundaries. To ensure consistency and reduce annotation errors, all segmentation masks were subsequently reviewed and confirmed by one independent clinical expert. This dual-review protocol was adopted to strengthen inter-observer reliability and to ensure accurate lesion localization, which is essential for downstream radiomics feature extraction. Cases with compromised boundary definition—caused by pleural effusion, bulky lymphadenopathy, atelectasis, extensive fibrosis, or other confounding pathology—were excluded from further analysis. This expert-driven, consensus-based protocol directly addresses KTA barriers of clinical skepticism (ii), adaptability to clinical complexity (iii), and reliability and quality challenges (xi) by ensuring that only high-quality, clinically validated masks advance to radiomics analysis.

### *C. Preprocessing and ROI-Based Cropping for Tumor-Focused Segmentation.*

All 3D CT volumes and corresponding masks were standardized using a multi-step preprocessing pipeline. CT scans were z-score normalized on a per-volume basis (mean/standard deviation), and masks were binarized to [0,1]. Volumes were resampled to $1 \times 1 \times 1$ mm³ isotropic spacing. During training, validation, and external testing, the ground-truth region of interest (ROI) center served as a simulated clinician click-point for extracting a fixed $64 \times 64 \times 64$ voxel crop. This standardized approach preserved peritumoral context and provided tolerance to moderate click-point deviations, provided the tumor remained within the bounding box. Cropping was performed with SimpleITK, with updated origins and complete logs maintained for reproducibility. This preprocessing enhanced signal-to-noise ratio, reduced irrelevant regions, and improved segmentation performance across all 3D DL models while preserving geometric fidelity [66]. This preprocessing and tumor-centered cropping strategy directly addresses KTA barriers of infrastructure and integration (i), adaptability to clinical complexity (iii), workflow disruption (vii), and reliability and quality challenges (xi) by standardizing inputs, reducing computational burden, and ensuring reproducible, tumor-focused segmentation across multi-center datasets.

### D. DL Segmentation Networks.

We selected 3D Attention U-Net [67], ResUNet [68], V-Net [69], ReconNet [68], and SAM-Med3D [70] for their complementary architectures, representing attention-based, residual, volumetric, lightweight, and foundation-model paradigms, respectively. Together, these architectures represent a broad spectrum of contemporary segmentation strategies. 3D Attention U-Net employs attention mechanisms for improved boundary localization, ResUNet leverages residual connections for stable optimization, and V-Net is a widely validated 3D model for volumetric CT imaging. ReconNet provides a lightweight and efficient alternative, while SAM-Med3D introduces recent advances in foundation models with strong generalization capabilities [68]. To ensure robustness, these networks were applied to both whole CT images and cropped (click-point–guided) tumor-centered volumes, enabling a direct evaluation of segmentation variability under different input strategies. Across the 12 datasets, approximately 72% of the cases were used for training, 18% for validation, and 10% for external testing. Specifically, cases from ten independent datasets, including LCTSC, LIDC-IDRI, Lung-Fused-CT-Pathology, NSCLC-Radiogenomics, NSCLC-Radiomics, QIN Lung CT, RIDER Lung CT, RIDER Pilot, SPIE-AAPM Lung CT Challenge, and TCGA-LUAD, were first randomly shuffled and subsequently partitioned into the training and validation cohorts, corresponding to approximately 72% and 18% of the total cases, respectively. Model selection was performed based on performance in the validation cohort. The two remaining independent datasets, LungCT-Diagnosis and NSCLC-Radiomics-Genomics, representing approximately 10% of the total cases, were kept separate from model development and used exclusively for external testing to evaluate model generalizability. Segmentation performance was measured using the Dice similarity coefficient, Intersection over Union (IoU), and Hausdorff distance, all computed via the AllMetrics library, which standardizes machine learning evaluation [71]. This was essential, as recent studies [72] have shown significant discrepancies in metric values across platforms (MATLAB, Python, R) and even across different Python libraries, raising concerns about reproducibility. Using AllMetrics ensured consistent, transparent, and comparable results. By selecting well-established networks and standardized evaluation, this step addresses KTA barriers of adaptability to clinical complexity (iii), reliability and quality challenges (xi), and lack of training or familiarity with AI tools (vi), while enhancing trust among clinical professionals.

### E. Radiomics Feature Extraction and Normalization.

All CT scans underwent rigorous quality control to exclude motion or reconstruction artifacts to ensure that observed performance differences are related to the model behavior rather than an image quality issue. Lung cancer lesions were manually segmented in 3D Slicer by two trained physicians and subsequently reviewed by one experienced thoracic medical expert for accuracy. A total of 497 radiomics features were extracted from each ROI, including both ground-truth annotations segmented by medical professionals and masks generated by the five DL networks, using the PySERA library within ViSERA 1.0, which is standardized in accordance with the Image Biomarker Standardization Initiative [73], [74]. These features comprised 50 first-order intensity features, including local intensity, intensity-based statistics, intensity histogram, and intensity-volume histogram features; 29 shape and morphological features; 408 texture features; and 10 moment-invariant features. The 408 texture features included 150 Gray-Level Co-occurrence Matrix (GLCM), 96 Gray-Level Run Length Matrix (GLRLM), 48 Gray-Level Size Zone Matrix (GLSZM), 48 Gray-Level Distance Zone Matrix (GLDZM), 15 Neighborhood Gray-Tone Difference Matrix (NGTDM), and 51 Neighboring Gray-Level Dependence Matrix (NGLDM) features. Gray-level discretization was performed using a fixed bin number (FBN) of 25 bins. Radiomics features were extracted from the original CT images without additional wavelet or Laplacian-of-Gaussian filtering. All features were normalized using min–max scaling to ensure comparability across datasets, prevent bias from differing feature ranges, and improve the stability and convergence of downstream classification analysis. Among the 12 datasets, only NSCLC-Radiogenomics, LungCT-Diagnosis, and NSCLC-Radiomics included documented OS data, comprising a total of 499 patients, while the remaining datasets were incorporated exclusively into the SSL framework or external testing. Both SL and SSL approaches were applied across multiple ML models to predict OS in patients with NSCLC. This step directly addresses KTA barriers of trust and accountability (iv), reliability and quality challenges (xi), adaptability to clinical complexity (iii), clinical distrust in AI decision-making (x), and workflow disruption/lack of standardization (vii), ensuring that radiomics analyses remain reproducible, interpretable, and clinically relevant.

### F. Statistical Radiomics Assessment.

Radiomics features extracted from DL-generated masks were compared to those from ground truth using Spearman correlation and paired t-tests to assess reproducibility, stability, and systematic bias due to segmentation variability. Radiomics captures tumor size, shape, higher-order textural patterns, and intratumoral heterogeneity, all of which are sensitive to segmentation differences. Ensuring that DL contours preserve these features is essential for maintaining prognostic reliability and preventing variability from impacting outcome models.

TABLE I
SUMMARY OF PATIENT, CLINICAL, IMAGING, AND OUTCOME CHARACTERISTICS ACROSS THE 12 TCIA DATASETS.

| Dataset | Included / Available Patients | Demographic / Clinical Characteristics Reported | Image Matrix | In-plane Pixel Spacing (mm) | Slice Thickness (mm) | OS Data Available | Prognostic Role | Ref. |
|---|---|---|---|---|---|---|---|---|
| **LCTSC** | 35 / 60 | Not reported | 512 × 512 × variable slices | 0.977 | 3 | No | Unlabeled cohort used in SSL | [53] |
| **LIDC-IDRI** | 289 / 1,010 | Not reported | 512 × 512 × variable slices | 0.6641–0.8223 | 1.25–2.5 | No | Unlabeled cohort used in SSL | [54] |
| **Lung-Fused-CT-Pathology** | 6/6 | 83.3% male; mean age: 74.8 ± 10.3 years; mean tumor size: 12.2 ± 1.9 mm | 512 × 512 × variable slices | 0.6602–0.7598 | 1.0–5.0 | No | Unlabeled cohort used in SSL | [55] |
| **LungCT-Diagnosis** | 49 / 61 | 59.6% aged ≥65 years; 53.2% male; predominantly stage III/IV | 512 × 512 × variable slices | 0.6230–0.8477 | 3.0–6.0 | Yes | External prognostic test cohort | [56] |
| **NSCLC-Radiogenomics** | 39 / 211 | Additional demographic characteristics not reported in the provided description | 512 × 512 × variable slices | 0.5488–0.7422 | 1.0–1.25 | Yes | External prognostic test cohort | [57] |
| **NSCLC-Radiomics** | 417 / 422 | 72.3% male; mean age: 66.2 ± 10.3 years; squamous cell carcinoma: 50.2%; large cell carcinoma: 30.1%; adenocarcinoma: 12.3%; NOS/NA: 7.4% | 512 × 512 × variable slices | 0.9766–0.9770 | 3 | Yes | Prognostic model-development cohort | [58] |
| **NSCLC-Radiomics-Genomics** | 53 / 89 | Not reported | 512 × 512 × variable slices | 0.6992–0.9766 | 5 | No | Unlabeled cohort used in SSL | [59] |
| **QIN Lung CT** | 38 / 47 | Not reported | 512 × 512 × variable slices | 0.6367–0.8359 | 3 | No | Unlabeled cohort used in SSL | [60], [61] |
| **RIDER Lung CT** | 12 / 32 | Not reported | 512 × 512 × variable slices | 0.7422–0.9180 | 1.25–5.0 | No | Unlabeled cohort used in SSL | [62] |
| **RIDER Pilot** | 4/8 | Not reported | 512 × 512 × variable slices | 0.7031–0.7578 | 5 | No | Unlabeled cohort used in SSL | [63] |
| **SPIE-AAPM Lung CT Challenge** | 44 / 77 | Not reported | 512 × 512 × variable slices | 0.5449–0.8477 | 1 | No | Unlabeled cohort used in SSL | [64] |
| **TCGA-LUAD** | 13 / 69 | Not reported | 512 × 512 × variable slices | 0.5703–0.6445 | 5.0–8.0 | No | Unlabeled cohort used in SSL | [65] |
| **Total** | **999 / 2,092** | Heterogeneous multicenter cohorts | **512 × 512 × variable slices** | — | — | **3 datasets with OS data** | SL/SSL prognostic analyses | — |

To evaluate normality, the Shapiro-Wilk test was applied, categorizing the data as "Normal" or "Not normal" based on the p-value. The Spearman correlation coefficient assessed the strength and direction of monotonic relationships between variables, while the Intraclass Correlation Coefficient (ICC) measured the consistency or reproducibility of the measurements. For non-normally distributed data, the Wilcoxon signed-rank test was used to compare paired samples. MANOVA was performed on multivariate data, with Wilks' lambda, Pillai's trace, Hotelling-Lawley trace, and Roy's greatest root reported to test group differences. The paired t-test was used for normally distributed data to identify significant variations between groups, reporting the t-value and the number of significant differences. To account for multiple hypothesis testing across the 497 radiomics features, feature-level p-values obtained from the paired t-test or Wilcoxon signed-rank test, as appropriate according to the normality assessment, were adjusted separately for each segmentation model using the Benjamini-Hochberg procedure to control the false discovery rate (FDR) at 0.05. Spearman correlation coefficients and ICC values were interpreted as measures of association and reproducibility, respectively, while MANOVA was treated as a global multivariate analysis; therefore, these analyses were not subjected to feature-wise multiple-comparison correction. To complement statistical significance testing, effect sizes were also reported where applicable. Cohen's $d_z$ was used for paired t-tests, and rank-biserial correlation was used for Wilcoxon signed-rank tests. Spearman correlation coefficients and ICC values were interpreted directly as measures of association and reproducibility, respectively.

Combining these methods allows for a rigorous evaluation of DL-based radiomics beyond correlation alone. This analysis directly addresses KTA barriers of trust and accountability (iv), variability and reliability (xi), workflow integration (vii), and clinician skepticism (ii). At the conclusion of this step, DL-generated masks are expected to show strong concordance with expert annotations. If discrepancies are found, further qualitative review and downstream classification or survival modeling will be performed to investigate and explain any variations. This framework strengthens the study's robustness and provides a generalizable model for assessing segmentation reliability across cancers and imaging modalities, advancing the integration of AI into clinical imaging practice.

### *G. Machine Learning Classification Analysis.*

SSL was included as a complementary downstream prognostic evaluation because outcome labels were available for only a subset of the imaging cohorts. Given prior evidence that SSL can provide better generalization than conventional SL when labeled data are limited, we used this framework to assess whether unlabeled imaging cases could improve prognostic modeling based on radiomics features derived from the different segmentation masks. We thus employed two complementary frameworks: SL and SSL.

In the SL framework, the labeled NSCLC-Radiomics cohort was randomly partitioned at the patient level into five non-overlapping folds. In each cross-validation iteration, four folds were used for model training, and the remaining fold was reserved for validation, such that each patient appeared in the validation set exactly once and no patient was shared between the training and validation subsets. For prognostic evaluation, NSCLC-Radiogenomics and LungCT-Diagnosis were used as external test cohorts and were held out from all prognostic model development, including classifier training, hyperparameter tuning, pseudo-label generation, validation, and model selection. Performance was evaluated using Accuracy, Precision, Recall, F1-score, Receiver Operating

Characteristic–Area Under the Curve (ROC-AUC), and Specificity, with results summarized as mean ± standard deviation across the five cross-validation folds and external evaluations.

In the SSL framework, NSCLC-Radiomics was likewise partitioned at the patient level into five mutually exclusive folds. For each iteration, four folds were used to train a logistic regression (LR) model, which was then used to generate pseudo-labels for the unlabeled development datasets, including LCTSC, LIDC-IDRI, Lung-Fused-CT-Pathology, QIN Lung CT, NSCLC-Radiomics-Genomics, RIDER Lung CT, RIDER Pilot, SPIE-AAPM Lung CT Challenge, and TCGA-LUAD. Pseudo-labels were assigned from the LR predictions and subsequently treated as training labels for the unlabeled cases. No additional confidence threshold or quality-based filtering was applied to exclude low-confidence pseudo-labels. The held-out NSCLC-Radiomics fold was excluded from both model training and pseudo-label generation and was used exclusively for validation, thereby preventing information leakage between the training and validation cohorts. The SSL model was subsequently retrained using the combined labeled training data and pseudo-labeled development data and evaluated on the held-out validation fold as well as the independent external test datasets. Throughout the prognostic analysis, the external test cohorts were maintained separately from all classifier training, pseudo-labeling, validation, hyperparameter tuning, and model-selection procedures.

To address the high dimensionality of radiomics features and reduce overfitting, dimensionality reduction strategies were applied, including feature selection algorithms (FSAs) and automated embedding algorithms (AEAs), totaling 38 approaches. FSAs covered filter-based methods (e.g., Chi-Square Test, Correlation Coefficient [CC], Mutual Information [MI], Information Gain (IG), Mutual Information Gain Ratio (MIGR)), statistical tests (e.g., Analysis of Variance [ANOVA] F-Test, Chi-Square P-value, Variance Thresholding [VT]), wrapper-based approaches (e.g., Recursive Feature Elimination [75], Sequential Forward Selection [SFS], Sequential Backward Selection [SBS]), embedded methods (e.g., Least Absolute Shrinkage and Selection Operator [Lasso], Elastic Net [ENet], Stability Selection), and ensemble strategies (e.g., Random Forest Importance [RF-Imp], Extra Trees [ETI], Permutation Importance [Perm-Imp]). Additional statistical controls, such as FDR, Family-Wise Error (FWE), and Variance Inflation Factor (VIF), were also incorporated.

The AEAs (19 methods) provided complementary transformations, including Principal Component Analysis (PCA), Kernel PCA (K-PCA), Independent Component Analysis (ICA/FastICA), Factor Analysis, Non-Negative Matrix Factorization (NMF), and nonlinear manifold learning methods such as t-distributed Stochastic Neighbor Embedding (t-SNE), Uniform Manifold Approximation and Projection (UMAP), Isomap, Locally Linear Embedding (LLE), Spectral Embedding, Multidimensional Scaling (MDS), and Diffusion Maps. Deep learning–based autoencoders (AE), as well as Feature Agglomeration, Truncated Singular Value Decomposition (SVD), and random projection techniques (Gaussian Random Projection (GRP) and Sparse Random Projection (SRP)), were also used for scalable dimensionality reduction. Each reduced feature set was evaluated with a comprehensive panel of 24 classifiers representing linear, probabilistic, kernel-based, tree-based, neural, and ensemble learning methods. These included Logistic Regression (LR), Linear Discriminant Analysis (LDA), Naïve Bayes (Gaussian [GNB], Bernoulli [BNB], Complement variants [CNB]), Support Vector Machines (SVMs) with multiple kernels, k-Nearest Neighbors (k-NN), Decision Trees (DT), Random Forest (RF), Extra Trees (ET), Gradient Boosting (GB), AdaBoost (AB), HistGradient Boosting (HGB), Light Gradient Boosting Machine (LGBM), Extreme Gradient Boosting (XGBoost), and Multi-Layer Perceptrons (MLP). Meta-ensembles such as Stacking, Voting (VC), and Bagging further enhanced predictive robustness. Additional classifiers included Nearest Centroid (NC), Gaussian Process Classifier (GPC), Decision Stump (DS), Dummy Classifier (DC), and Stochastic Gradient Descent Classifier (SGDC). Hyperparameters were optimized via five-fold cross-validation and grid search, with normalization, feature selection/embedding, and tuning restricted to fold-specific training data to prevent leakage into validation and external test sets.

By combining SL and SSL frameworks, dimensionality reduction, and a broad classifier panel validated on external datasets, this analysis ensured robustness, reproducibility, and interpretability. Importantly, this classification analysis also served as a direct evaluation of DL-generated masks compared with expert ground truth annotations. By testing whether radiomics features derived from DL masks achieve comparable or superior predictive performance relative to physician-delineated contours, the framework identified where DL masks performed better or worse than expert ground truth in both prediction and diagnostic tasks. Following this analysis, physicians with expertise in AI, radiomics, and ML-based classification reviewed the results to investigate why such differences occurred. This included examining what regions DL algorithms prioritized during segmentation, which prognostic features they captured, and what subtle information clinicians may have missed but DL models detected. Conversely, cases where DL performed worse were analyzed to reveal algorithmic limitations or overlooked tumor regions. Such physician-guided interpretation not only supports validation of DL outputs but also helps identify how much margin or peritumoral region may need to be consistently included for reliable diagnosis and prediction. In alignment with the KTA, this classification analysis addressed barriers of adaptability (iii), variability and reliability (xi), limitations in atypical or complex cases (ix), and workforce resistance (viii) by demonstrating that DL models can complement rather than replace clinical expertise.

### H. Qualitative review Analysis.

Six physicians participated in the blinded qualitative evaluation of AI-generated masks, without knowledge of model identity or quantitative performance results. Three of these physicians also contributed to ground-truth lesion delineation and verification, while three additional external physicians participated only in the qualitative evaluation. All six physicians independently completed the physician-assessment questionnaire used for the qualitative analysis. Physician involvement therefore spanned ground-truth delineation and verification, assessment of AI-generated segmentations, and interpretation of radiomics, prognostic, trust, and workflow-related findings. However, clinician feedback was not iteratively incorporated into model retraining or inference in the present retrospective study. The evaluation involved comparing the outputs of five DL models to the ground truth and extracting radiomic features from ROIs for survival prediction tasks. This analysis summarizes key evaluation areas, including tumor delineation, spatial differences, and the clinical relevance of AI mask deviations for prognosis and survival prediction. The evaluation protocol spanned seven domains with 21 questions, as listed in Supplemental File 1. In (q1), radiologists judged whether AI masks define tumor volumes as more clinically meaningful than ground truth. (q2) asked if AI–ground truth differences were concentrated in the peritumoral zone, while (q3) evaluated the clinical meaningfulness or justification of those differences. (q4) assessed preservation of internal tumor heterogeneity, and (q5) focused on boundary quality. Prognostic value was examined in (q6), asking whether AI–ground truth differences captured survival-relevant tissue, and in (q7), whether deviations in tumor volume were harmful or beneficial for prediction. Trust and preference were probed in (q8), addressing the likelihood that AI masks could outperform ground truth in subtle cases, and in (q9), whether experts would prefer AI masks for survival prediction tasks. Overall assessments included (q10) diagnostic success, (q11) prognostic success, (q12) confidence in evaluation, (q13) sufficiency of geometric metrics, and (q14) identification of additional evaluation needs. Workflow and adoption were explored in (q15) usability, (q16) clinical actionability, (q17) robustness, (q18) time-saving potential, (q19) trust in high-stakes clinical tasks, and (q20) the value of confidence maps. Finally, (q21) captured expert opinion on the broader role of AI in clinical practice—whether to replace physicians, provide initial masks for refinement, or act as a supportive tool. Differences among the five segmentation models for selected physician ratings (q10, q11, q13, and overall ratings) were assessed using the Friedman test. Kendall's W was reported as the corresponding effect-size measure. The remaining questionnaire items were analyzed descriptively. By embedding physician expertise in this way, the analysis strengthens transparency, clinical trust, and interpretability, addressing KTA barriers of clinical skepticism (ii), distrust in AI decision-making (x), and workforce resistance.

## III. RESULTS

### A. Lesion Segmentation Results.

As shown in Fig. 1, V-Net on cropped (click-point) images achieved the highest Dice and IoU values among the evaluated models, while SAM-Med3D achieved the lowest Hausdorff distance in the test set. In the test dataset, VNet achieves the highest values for both metrics (Dice: 0.83 ± 0.07, IoU: 0.71 ± 0.09), with a relatively low Hausdorff distance (10.04 ± 3.73). In the validation set, VNet also shows top performance (Dice: 0.77 ± 0.15, IoU: 0.64 ± 0.16, Hausdorff Distance: 11.66 ± 5.50). The difference in testing performance between VNet and the other networks is statistically significant ($p<0.05$, paired t-test). ReconNet performs well in the test set with a Dice score of 0.79 ± 0.11 and IoU of 0.67 ± 0.13, showing a Hausdorff Distance of 11.39 ± 3.96. In the validation set, it slightly underperforms (Dice: 0.72 ± 0.17, IoU: 0.59 ± 0.18, Hausdorff Distance: 13.17 ± 5.42). ResUNet performs moderately in both datasets. In the test set, it shows a Dice score of 0.75 ± 0.09 and an IoU of 0.61 ± 0.10, with a Hausdorff Distance of 19.91 ± 6.91. In the validation set, its performance drops slightly (Dice: 0.74 ± 0.10, IoU: 0.60 ± 0.12, Hausdorff Distance: 20.17 ± 7.73). 3D Attention U-Net, though providing competitive performance in the test set (Dice: 0.77 ± 0.08, IoU: 0.64 ± 0.10), exhibits a significantly higher Hausdorff Distance (33.61 ± 4.10) compared to the other models, indicating more variability in boundary predictions. In the validation set, its performance remains consistent (Dice: 0.76 ± 0.11, IoU: 0.63 ± 0.13, Hausdorff Distance: 33.48 ± 4.40). SAM-Med3D demonstrates the lowest performance in both Dice score and IoU, particularly in the validation set (Dice: 0.61 ± 0.27, IoU: 0.49 ± 0.24). In the test set, SAM-Med3D achieved the lowest test-set Hausdorff distance (4.13 ± 1.55), despite lower Dice and IoU values. Moreover, all networks trained on whole images underperformed compared to cropped images, yielding significantly lower results (Dice: 0.30, IoU: 0.20, Hausdorff Distance: 9; $p<<0.001$, paired t-test with Benjamini–Hochberg FDR correction). VNet emerges as the most robust model across testing and validation datasets, combining high accuracy (Dice and IoU) with low boundary variability (Hausdorff Distance), with superior performance compared to the other networks. The hyperparameters and architectural characteristics of the DL segmentation networks are provided in Supplemental File 2.

### B. Statistical Radiomics Assessment Results

As shown in Table II, the statistical analysis of radiomics features extracted from the DL-generated masks revealed strong performance across all evaluated models, with V-Net demonstrating the most consistent results. The Spearman correlation coefficients (ρ) demonstrated consistent monotonic relationships between the DL-derived and ground-truth radiomics features, ranging from 0.72 to 0.76, with VNet, ResUNet, and 3D Attention U-Net sharing the highest average correlation value of 0.76. The ICC values, which assessed reproducibility, ranged from 0.61 to 0.67, with V-Net showing a high ICC of 0.65, reflecting moderate to high consistency. For non-normally distributed features, differences between DL-derived and ground-truth radiomics measurements were assessed using the Wilcoxon signed-rank test, with statistical significance determined using Benjamini-Hochberg FDR-adjusted p-values. Significant feature-level differences were observed across all five segmentation models. MANOVA further demonstrated multivariate differences between DL-derived and ground-truth radiomics measurements across all models, with Wilks' λ, Pillai's trace, Hotelling-Lawley trace, and Roy's greatest root summarized in Table II. Notably, 3D Attention U-Net also demonstrated radiomics measurements broadly comparable to those of VNet across several evaluated metrics. Feature-level comparisons were performed using paired t-tests for normally distributed features and Wilcoxon signed-rank tests for non-normally distributed features. After Benjamini-Hochberg FDR correction, the number of features showing significant differences between DL-derived and ground-truth radiomics measurements ranged from 26 to 97 out of 497 features. VNet and 3D Attention U-Net exhibited the fewest significant feature-level differences (36/497 and 26/497, respectively; adjusted $p<0.05$). To complement statistical significance testing, effect sizes were quantified using Cohen's $d_z$ for paired t-tests and rank-biserial correlation for Wilcoxon signed-rank tests. Complete feature-level statistical results, including FDR-adjusted p-values and effect-size estimates, are provided in Supplemental File 3.

### C. Survival Prediction Analysis

In 5-fold cross-validation (Fig. 2), SSL consistently outperformed SL, achieving higher accuracy, recall, and F1-scores with tighter deviations, while SL showed more variability. For instance, SSL with MI–MLP–ResUNet reached 0.88 ± 0.003 accuracy, 0.88 ± 0.003 recall, and 0.83 ± 0.004 F1, compared to 0.75 ± 0.005, 0.75 ± 0.005, and 0.64 ± 0.007 for SL, with similar advantages observed for MIGR–ET–ground truth and ReliefF–GB–SAM-Med3D. The LASSO–LGBM–VNet configuration stood out, consistently achieving the strongest validation results (0.88 ± 0.003 accuracy, 0.83 ± 0.004 F1, and 0.65 ± 0.031 AUC), confirming the power of VNet as a segmentation backbone. Independent test evaluations supported these findings, though SSL–SL gaps were narrower: while models like MI–MLP–ResUNet and ReliefF–GB–SAM-Med3D showed similar performances across paradigms, VNet again demonstrated clear superiority. Under SSL, LASSO–LGBM–VNet achieved the highest test accuracy (0.88 ± 0) and a strong AUC (0.76 ± 0.048), whereas its SL counterpart, despite maintaining accuracy (0.87 ± 0.012), dropped sharply in AUC (0.40 ± 0.058) ($p<0.05$, paired t-test with Benjamini-Hochberg FDR correction). Likewise, GRP–VC–3D Attention further highlighted SSL's advantage, achieving 0.81 ± 0.025 accuracy and 0.82 ± 0.029 F1, while SL lagged at 0.72 ± 0.079 accuracy and 0.75 ± 0.062 F1.

Taken together, the cross-validation and independent test results demonstrate that SSL provides significantly more stable and reliable predictive performance compared to SL. SSL models consistently achieved higher accuracy, recall, and F1-scores while also sustaining stronger AUC and specificity, particularly in configurations involving LASSO–LGBM–VNet and MIGR–ET–ground truth. Importantly, VNet stood out as the most powerful backbone among the segmentation models, delivering superior performance across both validation and test phases and showing greater robustness in SSL than in SL. While SL occasionally approached SSL on the test set (e.g., ReliefF–GB–SAM-Med3D and UMAP–AB–ReconNet), it was generally associated with larger standard errors and less consistent generalization. These findings collectively highlight both the overall robustness of SSL and the particularly strong role of VNet in driving state-of-the-art classification outcomes. The hyperparameters, architectural characteristics, selected features, and results of five-fold cross-validation and external tests for the FSAs/AEAs combined with classifiers on each dataset, using radiomics features extracted from DL-generated masks and ground truth, are provided in Supplemental Files 4–15.

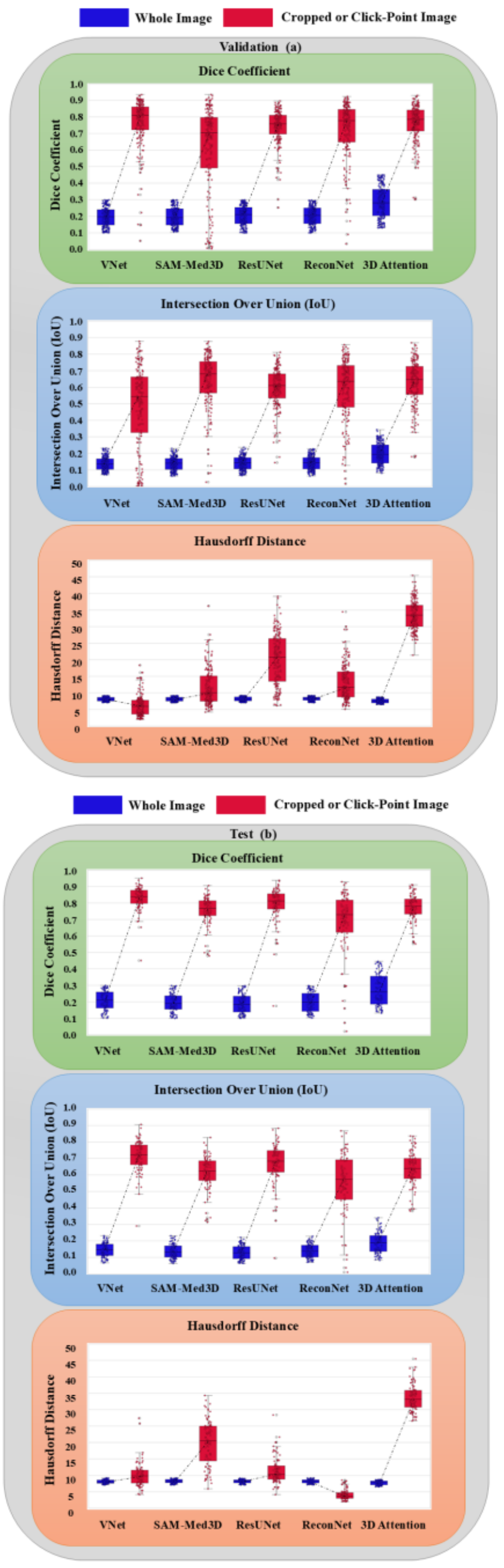

**Fig. 1.** Comparative performance of five lung lesion segmentation models on whole (blue) vs. cropped (red) CT images. (a) shows the results on the validation sets, while (b) presents the results on the test sets. The five segmentation networks evaluated include 3D Attention U-Net, ResUNet, V-Net, ReconNet, and SAM-Med3D. Three evaluation metrics were used to compare the models: Dice Similarity Coefficient, Intersection over Union (IoU), and Hausdorff Distance.

### *D. Qualitative Visual Assessment*

Six board-certified thoracic radiologists (5–15 years of experience) independently evaluated segmentation masks from five models. The heatmap in Table III summarizes average Likert ratings across 19 scored questions (q1–q13, q15–q20), with q14 excluded as descriptive and q21 treated as categorical. For most items, higher scores indicated more favorable outcomes, except for q5 (boundary noisiness), where lower scores were preferable. For q21 (Role of AI in clinical practice), radiologists selected among three options: Option A — Replace physicians entirely, Option B — Provide an initial mask for physicians to refine, or Option C — Assist physicians as a support/check tool.

Section 1 — Overall Impression (q1): Ratings were modest (≈2.2–2.8), indicating that AI masks were not consistently judged more clinically meaningful than ground truth. (Descriptive only).

Section 2 — Spatial Differences (q2–q3): Peritumoral concentration (q2, higher = better) scored highest for 3DAttention and VNet (≈3.8), while the clinical meaningfulness of differences (q3, higher = better) also favored these models (≈3.5) compared to SAM-Med3D (≈1.5). (Descriptive only).

Section 3 — Tumor Heterogeneity & Boundary Quality (q4–q5): Preservation of heterogeneity (q4, higher = better) ranged from approximately 1.8 to 3.2. Boundary noisiness (q5, lower = better) was lowest for VNet (1.8), indicating smoother physician-perceived boundaries, whereas 3D Attention U-Net received the highest noisiness rating (3.5). (Descriptive only).

Section 4 — Survival Prediction Relevance (q6–q7): Scores were mid-range (≈1.8–3.0, higher = better), suggesting that AI–ground truth deviations were somewhat prognostically relevant but not strongly impactful. (Descriptive only).

Section 5 — Trust & Preference (q8–q9): Ratings were modest (≈1.7–2.7, higher = better), reflecting limited clinician preference for adopting AI masks over ground truth in survival prediction tasks. (Descriptive only).

Section 6 — Diagnostic success (q10) ranged from approximately 2.5 to 3.5, while prognostic success (q11) ranged from approximately 1.8 to 3.4. Friedman tests revealed no statistically significant differences among the five models in diagnosis-oriented evaluation (q10: $\chi^2$ = 2.140, p = 0.710, Kendall's W=0.089) or prediction-oriented evaluation (q11: $\chi^2$ = 3.802, p = 0.433, Kendall's W=0.190). Overall confidence (q12, higher = better) reached some of the strongest values (up to 4.0). Sufficiency of geometric metrics (q13, higher = better) was consistently low (≈2.2–2.3), and the Friedman test again confirmed no significant differences among models ($\chi^2$ = 1.714, p = 0.788, Kendall's W=0.071). A global Friedman test of overall ratings also showed no significant differences ($\chi^2$ = 4.205, p = 0.379, Kendall's W=0.175). For q14, physicians agreed that trustworthy AI segmentation requires evaluations beyond geometric accuracy, emphasizing clinical review, radiomics stability, survival prediction, workflow impact, and robustness to noise. They considered these more effective than Dice or IoU alone, as they better connect segmentation to clinical outcomes, usability, and real-world trust.

Section 7 — Workflow & Adoption (q15–q20, q21): Usability (q15) and clinical actionability (q16) received moderate ratings, while robustness (q17) ranged from approximately 2.3 to 3.3. Time-saving potential (q18) was rated relatively highly (≈3.3–4.3). In contrast, trust in using AI masks for high-stakes tasks without physician refinement (q19) was consistently low (≈1.0–1.8). Interpretability via confidence maps (q20) received high ratings among the available responses (≈4.2–4.4), emphasizing the importance of transparency for clinical acceptance. For q21 (categorical), radiologists indicated that AI should provide an initial mask for physician refinement and/or serve as a supportive tool rather than replace physicians.

TABLE II
STATISTICAL ANALYSES OF RADIOMICS FEATURES EXTRACTED FROM THE DL-GENERATED MASKS, COMPARED TO THE GROUND TRUTH (GT). ICC: INTRACLASS CORRELATION COEFFICIENT.

| Method | Test Type | ReconNet | ResUNet | VNet | 3D Attention | SAM-Med3D |
|---|---|---|---|---|---|---|
| **Shapiro (Normal/Not -normal)** | Ground Truth (314/183) | 331/166 | 282/215 | 364/133 | 298/199 | 340/157 |
| **Spearman** | Correlation Coefficient | 0.72 | 0.76 | 0.76 | 0.76 | 0.74 |
| **ICC** | Coefficient | 0.62 | 0.65 | 0.65 | 0.61 | 0.67 |
| **Wilcoxon** | W Statistic | 94.21 | 123.01 | 89.71 | 69.83 | 105.52 |
| **MANOVA** | Wilks' λ | 0.04 | 0.02 | 0.03 | 0.02 | 0.04 |
| | Pillai's Trace | 0.96 | 0.98 | 0.97 | 0.98 | 0.96 |
| | Hotelling-Lawley Trace | 21.59 | 57.23 | 37.00 | 58.57 | 21.47 |
| | Roy's Greatest Root | 21.59 | 57.23 | 37.00 | 58.57 | 21.47 |
| | Decision | Rejected | Rejected | Rejected | Rejected | Rejected |
| **Feature-level comparison** | Number of features with significant differences after Benjamini-Hochberg FDR correction (adjusted p<0.05) / 497 total features | 83/497 | 97/497 | 36/497 | 26/497 | 49/497 |

**Note:** Feature-level effect sizes were quantified using Cohen's $d_z$ for paired t-tests and rank-biserial correlation for Wilcoxon signed-rank tests. Complete effect-size results are provided in Supplemental File 3.

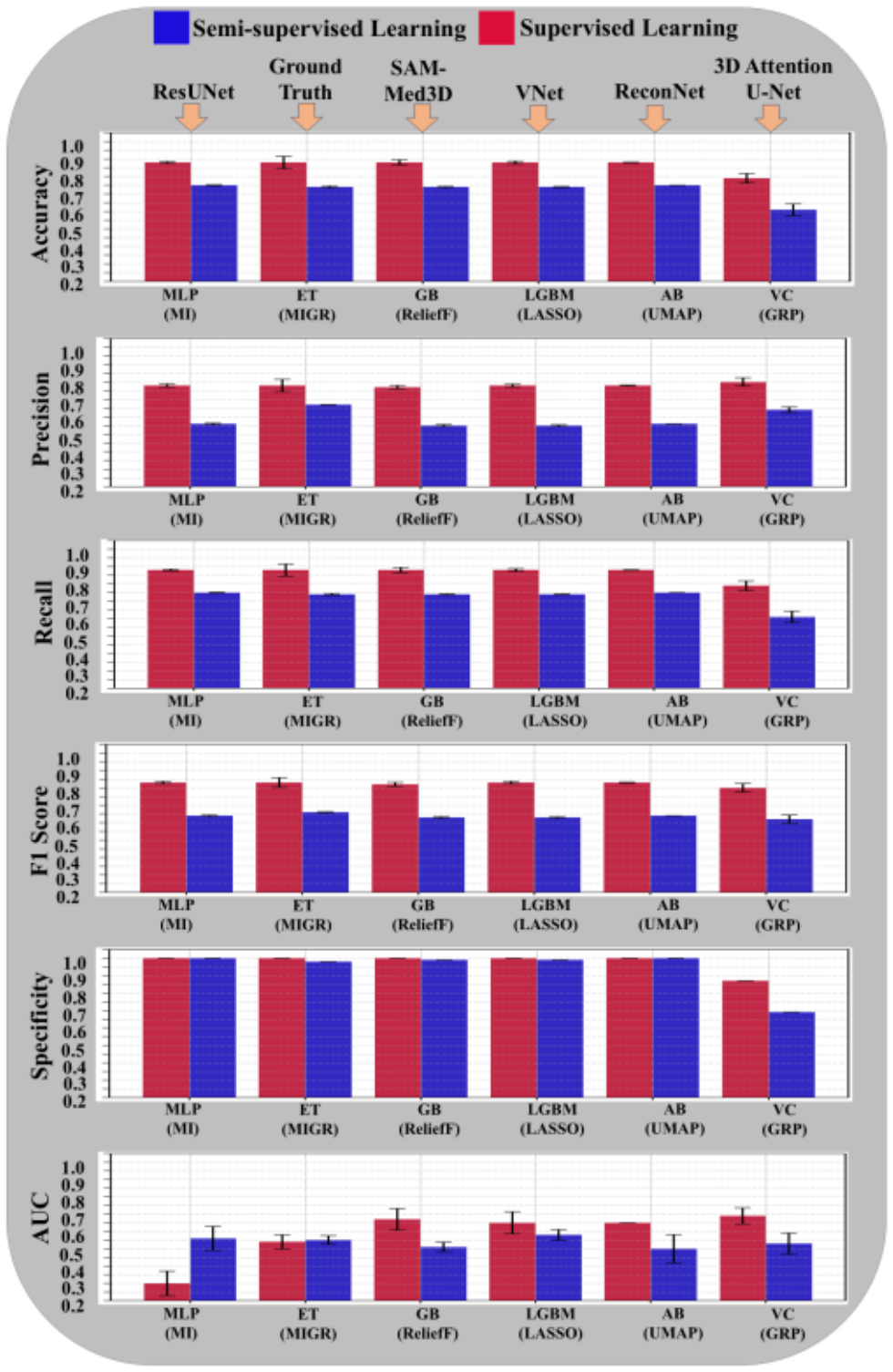

**Fig. 2**. Five-fold cross-validation performance of classification pipelines based on radiomics features extracted from segmentation masks generated by five models and Ground-Truth masks. Red bars correspond to semi-supervised learning and blue bars to supervised learning, with error bars showing standard deviation across folds.

TABLE III.
HEATMAP OF QUALITATIVE EVALUATION RESULTS DERIVED FROM PHYSICIANS FOR MASKS SEGMENTED BY DEEP LEARNING.

| Question | 3DAttention | SAM-Med3D | ReconNet | ResUNet | VNet |
|---|---|---|---|---|---|
| q1 | 2.8 | 2.2 | 2.2 | 2.3 | 2.7 |
| q2 | 3.8 | 2 | 2.7 | 2.8 | 3.8 |
| q3 | 3.5 | 1.5 | 3 | 2.5 | 3.5 |
| q4 | 3.2 | 1.8 | 2.7 | 2.8 | 3 |
| q5 | 3.5 | 2 | 2.5 | 3 | 1.8 |
| q6 | 2.8 | 1.8 | 2.3 | 2.3 | 2.6 |
| q7 | 3 | 2.7 | 2.2 | 2.3 | 2.7 |
| q8 | 2.7 | 1.8 | 2 | 2.2 | 3 |
| q9 | 2.3 | 1.7 | 2 | 1.5 | 2.7 |
| q10 | 3.5 | 2.5 | 3.3 | 2.8 | 3.3 |
| q11 | 2.6 | 1.8 | 2.4 | 2.2 | 3.4 |
| q12 | 3.7 | 4 | 3.7 | 3.8 | 3.5 |
| q13 | 2.3 | 2.3 | 2.3 | 2.2 | 2.2 |
| q15 | 3.2 | 2.5 | 3.3 | 2.7 | 3.3 |
| q16 | 3.7 | 3.5 | 2.8 | 3.2 | 3.7 |
| q17 | 3.3 | 2.7 | 3 | 2.3 | 3.2 |
| q18 | 4.2 | 3.3 | 4.2 | 3.5 | 4.3 |
| q19 | 1.7 | 1 | 1.7 | 1 | 1.8 |
| Q20 | 4.2 | 4.2 | 4.2 | 4.2 | 4.4 |

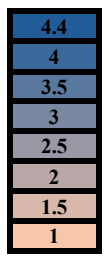

Friedman tests were applied to q10, q11, q13, and the overall ratings, with Kendall's reported as the corresponding effect-size measure. All tests were non-significant (p>0.3), with small effect sizes (W=0.071–0.190), indicating limited differences among the five models in these physician-rated domains. The remaining Likert-rated questions (q1–q9, q12, and q15–q20) were analyzed descriptively, while q14 was treated as an open-ended response and q21 as a categorical response. Their distributions showed generally consistent patterns across models, with no evidence of marked differences in physician-rated performance. Collectively, these findings indicate limited physician-rated differences among the five models in the evaluated clinical domains despite differences in technical segmentation performance. Qualitative assessments are provided in Supplemental File 16.

## IV. DISCUSSION

Our results show that geometric segmentation performance alone does not fully correspond to radiomics reproducibility, downstream prognostic performance, or physician assessment, supporting a multidimensional evaluation of AI-generated tumor segmentations. Among the five well-known models, VNet achieved the best testing Dice (0.83) and strong radiomics reproducibility, while SSL with VNet–LASSO–LGBM reached 0.88 accuracy, surpassing SL and underscoring the utility of unlabeled data. SAM-Med3D showed lower Dice and IoU performance than the task-specific segmentation networks, although it achieved a relatively low Hausdorff distance in the test set. Several factors may explain this pattern. First, SAM-Med3D is a general-purpose foundation model trained across heterogeneous medical imaging domains, whereas architectures such as VNet were optimized specifically for volumetric segmentation and may better capture lung tumor morphology in CT. Second, foundation-model performance can be sensitive to the prompting strategy, including the type, location, and number of prompts used for lesion localization. Third, domain differences between the data used for SAM-Med3D pretraining and the heterogeneous lung cancer CT cohorts evaluated here may have introduced a degree of domain shift. Additional task-specific fine-tuning, optimized prompting strategies, or adaptation to lung cancer CT could therefore improve its performance. Accordingly, our findings should not be interpreted as indicating that foundation models are inherently inferior, but rather that, under the conditions evaluated in this study, the task-specific networks achieved better overall overlap performance. ReconNet, despite weaker geometric scores, produced radiomics features with notable prognostic value, and 3D Attention U-Net captured peritumoral heterogeneity most effectively. Still, Friedman tests showed no significant differences for q10 (diagnostic success), q11 (prognostic success), or q13 (sufficiency of geometric metrics), with small Kendall's W effect sizes, indicating limited differences among the models in these physician-rated domains. The novel contribution of this work lies in its integrated evaluation framework, which unites geometric metrics, radiomics reproducibility, survival modeling, and physician-centered assessments under the KTA paradigm. While this study focused on handcrafted radiomics because these features provide predefined and interpretable descriptors of tumor intensity, shape, and texture, AI-driven feature representations are increasingly used in predictive modeling. Deep features learned directly from medical images may capture complementary patterns that are not fully represented by conventional radiomics. Future studies should investigate whether segmentation variability similarly affects learned imaging features and whether combining handcrafted radiomics with deep features can further improve prognostic performance and generalizability [76], [77]. We demonstrate that (i) SSL pipelines with VNet can enhance prognostic prediction, (ii) clinical reliability cannot be judged from geometric scores alone but requires outcome-linked metrics and physician input, and (iii) DL segmentation should act as an assistive tool rather than a replacement for clinicians. Physicians further highlighted that several AI-generated masks exhibited noisy or unstable boundaries rather than consistently biologically meaningful differences, which may limit prognostic stability. However, boundary quality varied across models, with VNet receiving the lowest boundary-noisiness rating (q5), indicating comparatively smoother physician-perceived contours. They also noted that confidence maps may improve transparency, although robust segmentation remains essential. Physicians' qualitative evaluations provided essential explanatory context for the quantitative findings. AI-generated masks were occasionally judged to be less clinically meaningful than the ground truth, primarily due to inconsistent and noisy boundaries, challenges at spiculated and craniocaudal margins, and geometric instability. These discrepancies seldom corresponded to biologically relevant regions but rather reflected random noise, which likely contributed to the inferior prognostic performance of AI masks compared to ground truth. While AI sometimes captured broader peritumoral regions or additional structures not included in ground truth masks, this was inconsistent and often contaminated with non-tumoral tissue, reducing reliability. Physicians also noted that ground truth segmentations, while occasionally biased by high-contrast thresholds, remained cleaner and more trustworthy, emphasizing that stability and interpretability often outweigh minor geometric precision in clinical settings. Collectively, these evaluations confirm that current AI delineations introduce instability rather than reliable new information, limiting their direct clinical value.

Within the KTA framework, these findings relate to several of the 11 barriers to AI deployment identified in our previous KTA survey study [2]. Section 1 (Overall Impression) reflected clinical skepticism (ii), as physicians judged that AI masks were rarely more meaningful than ground truth, and misregistration undermined confidence. Section 2 (Spatial Differences) addressed adaptability (iii): while VNet and 3D Attention occasionally captured broader peritumoral areas—known to carry prognostic significance—physicians stressed that these regions were captured inconsistently. Section 3 (Tumor Heterogeneity & Boundaries) engaged usability and reliability (vii, xi). Although AI masks sometimes preserve heterogeneity by capturing larger ROIs, this often came at the cost of including irrelevant tissue. Fragmented boundaries resulted in high Hausdorff distances and unstable radiomics features, whereas ground truth contours provided smoother, more reproducible segmentations.

Section 4 (Survival Prediction) aligned with outcome relevance (x). Physicians consistently linked noisy AI boundaries to weaker prognostic stability, suggesting that unstable tumor representations may have contributed to reduced prognostic performance. While SSL pipelines improved generalization and reduced annotation burden, their success was still dependent on stable segmentation quality. Section 5 (Trust & Preference) highlighted the confidence barrier (iv). Radiologists emphasized that interpretability tools, such as confidence or uncertainty maps, would improve transparency and trust, but they could not compensate for fundamental segmentation errors. Section 6 (Summaries & Assessment) related to standardization (vii). Physicians explicitly rejected Dice/IoU as sufficient measures of trustworthiness, explaining that geometric scores may misrepresent clinical value: models with high scores could still fail at margins, while those with modest scores occasionally captured useful prognostic regions.

Section 7 (Workflow & Adoption) addressed workflow disruption and workforce resistance (vii, viii). Physicians reported that AI masks often increased workload, since noisy and inconsistent boundaries required extensive correction. In some cases, it was easier to redraw from scratch than to refine the AI output. This inefficiency limited adoption despite the technical promise. Moreover, for high-stakes tasks such as radiotherapy planning, even small errors in tumor boundary delineation could compromise staging, surgical margins, or dose delivery. Physicians indicated that AI could potentially reduce interobserver variability and serve as a helpful starting point, but strongly emphasized that its role should remain assistive—providing an initial mask for refinement—rather than autonomous. This mitigates concerns about oversight and job displacement while reinforcing the importance of physician supervision in clinical workflows.

When integrating these findings, VNet emerges as a promising candidate for further clinical evaluation based on its segmentation accuracy, radiomics reproducibility, and prognostic performance. Descriptive physician assessments also indicated smoother boundaries, favorable peritumoral representation, and potential workflow benefits. However, Friedman tests revealed no statistically significant differences among models in the selected physician-rated domains. Thus, VNet's potential for clinical translation reflects an overall interpretation of the combined quantitative and descriptive findings rather than a statistically demonstrated clinician preference. From a KTA perspective, these findings suggest its potential to address barriers related to infrastructure and integration (i), adaptability (iii), clinical skepticism (ii), trust (iv), reliability (xi), and workflow disruption (vii). Its performance within SSL pipelines further supports its potential for clinical evaluation, although prospective validation is needed to establish its clinical utility.

Despite these contributions, several limitations remain. Prognostic validation was limited to three datasets, and the expert evaluation included only six medical professionals. Therefore, the physician assessment should be considered an initial evaluation of clinical perception and acceptance rather than a definitive representation of broader clinical practice. In addition, Likert ratings sometimes exhibited ceiling effects, and the qualitative assessments were inherently interpretive. These limitations highlight the need for larger, multicenter studies involving a more diverse group of clinicians to validate the generalizability, usability, and clinical acceptance of the proposed framework. Ethical, legal, financial, and privacy challenges (v) were not addressed, representing the major unresolved KTA barrier. As a practical next step, we plan to integrate eye-tracking systems into click-point segmentation to enhance usability and reduce user burden, thereby making interactions more intuitive. We will also pilot VNet-based segmentation within real-world clinical workflows, enabling prospective evaluation of usability, physician-AI collaboration, and prognostic value in practice. Larger, multi-institutional studies will be necessary to confirm generalizability, quantify the longitudinal benefits of outcomes, and address the remaining ethical and regulatory challenges. Moreover, a previous study [78] has reported that several wavelet-transformed radiomics features are sensitive to segmentation variability. Future studies should therefore investigate whether the effects observed here differ across image-filtering strategies. Because no confidence-based filtering was applied to pseudo-labels, future studies should investigate probability-thresholding or uncertainty-based selection strategies to reduce potential noise from low-confidence pseudo-labels. Although clinicians contributed across multiple stages of the study, the current framework did not include a closed iterative feedback loop in which physician corrections were used to update the AI model. Future work will incorporate active learning and iterative clinician feedback to enable a fully interactive clinician-guided/evaluated system. Finally, although the fixed 64 × 64 × 64 voxel bounding box accommodates all tumors in our dataset, larger lesions may be incompletely captured, consistent with reported variability in lung tumor volumes [79]. Future studies will investigate adaptive cropping strategies to address this limitation.

## V. Conclusion

Our study shows that although segmentation models differed in geometric accuracy, radiomics reproducibility, and prognostic performance, physician-based assessments revealed only limited differences among the models across several evaluated clinical domains. VNet achieved the highest Dice (0.83 ± 0.07) and IoU (0.71 ± 0.09) scores among the evaluated models, with a Hausdorff distance of 10.04 ± 3.73, strong radiomics reproducibility (Spearman correlation = 0.76, ICC = 0.65), and prognostic performance under SSL (accuracy = 0.88 ± 0.003, F1 = 0.83 ± 0.004, specificity = 1.0). Physician assessments showed no significant differences in diagnostic outcomes, but VNet excelled in segmentation, radiomics, and predictive performance, making it favorable for clinical workflows. This study addressed 10 of the 11 barriers to AI adoption in medical imaging, demonstrating that technical reproducibility, physician oversight, and workflow alignment are critical for trusted clinical implementation. Importantly, clinicians prefer AI as a tool for refinement, not replacement. We propose a clinical software platform integrating VNet-based segmentation with click-point and gaze-tracking, paired with active learning and SSL-derived radiomics, to streamline segmentation, enhance trust, and integrate seamlessly into clinical workflows.

## Acknowledgment

This study was supported by the Virtual Collaboration Group and the Technological Virtual Collaboration (TECVICO CORP.) based in Vancouver, BC, Canada.